\documentclass[11pt,a4paper]{article}
\usepackage[hyperref]{emnlp2020}
\usepackage{times}
\usepackage{latexsym}

\usepackage{todonotes}

\usepackage{amsmath,amssymb,bm}
\usepackage{graphicx}
\usepackage{multirow}

\usepackage{comment}

\aclfinalcopy

\usepackage{microtype}

\def\aclpaperid{1046} 

\DeclareMathOperator*{\E}{\mathbb{E}}

\title{Local Additivity Based Data Augmentation for Semi-supervised NER}

\author{
 Jiaao Chen\thanks{\ \ Equal contribution.} , Zhenghui Wang$^*$, Ran Tian$^1$, Zichao Yang$^2$, Diyi Yang \\
 Georgia Institute of Technology, $^1$ASIT Japan / Google, $^2$Citadel Securities\\
\texttt{\{jchen896,zhwang,dyang888\}@gatech.edu} \\
}

\date{}

\begin{document}
\maketitle
\begin{abstract}
Named Entity Recognition (NER) is one of the first stages in deep language understanding yet current NER models heavily rely on human-annotated data. In this work, to alleviate the dependence on labeled data, we propose a 
\textbf{L}ocal \textbf{A}dditivity based \textbf{D}ata \textbf{A}ugmentation (LADA) method for semi-supervised 
NER, in which we create virtual samples by interpolating sequences {\em close} to each other. Our approach
has two variations: Intra-LADA and Inter-LADA, where Intra-LADA performs interpolations among tokens within one sentence, and Inter-LADA samples different sentences to interpolate.
Through linear additions between sampled training data, LADA creates an infinite amount of labeled data
and improves both entity and context learning.
We further extend LADA to the semi-supervised setting by designing a novel consistency loss for unlabeled data. Experiments conducted on two NER benchmarks demonstrate the effectiveness of our methods over several strong baselines. We
have publicly released our code at \url{https://github.com/GT-SALT/LADA}.
\end{abstract}

\section{Introduction}
Named Entity Recognition (NER) that aims to detect the semantic category of entities (e.g., persons, locations, organizations) in unstructured text \cite{ner-sekine2007},  is an essential prerequisite for many NLP applications. Being one of the most fundamental and classic sequence labeling tasks in NLP, there have been
extensive research from traditional statistical models like Hidden Markov Models \cite{zhou-su-2002-named} and Conditional Random Fields  \cite{10.5555/645530.655813}, to neural network based models such as LSTM-CRF \cite{lample-etal-2016-neural} and BLSTM-CNN-CRF \cite{DBLP:journals/corr/MaH16}, and to recent pre-training and fine-tuning methods like ELMO \cite{DBLP:journals/corr/abs-1802-05365}, Flair \cite{akbik-etal-2018-contextual} and BERT \cite{devlin-etal-2019-bert}. However, most of those models still heavily rely on \emph{abundant annotated data} to yield the state-of-the-art results \cite{lin2020triggerner}, making them hard to be applied into new domains (e.g., social media, medical context or low-resourced languages) that lack labeled data.

Different kinds of data augmentation approaches have been designed to alleviate the dependency on labeled data for many NLP tasks, and can be categorized into two broad classes: 
(1) adversarial attacks at token-levels such as word substitutions \cite{kobayashi-2018-contextual, DBLP:journals/corr/abs-1901-11196} or adding noise \cite{lakshmi-narayan-etal-2019-exploration}, (2) paraphrasing at sentence-levels such as back translations \cite{DBLP:journals/corr/abs-1904-12848} or submodular optimized models  \cite{kumar-etal-2019-submodular}. 
The former has already been used for NER but struggles to create diverse augmented samples with very few word replacements.
Despite being widely utilized in many NLP tasks like text classification, the latter often fails to maintain the labels at the token-level in those paraphrased sentences, thus making it difficult to be applied to NER.

We focus on another type of data augmentations called mixup~\cite{DBLP:journals/corr/abs-1710-09412}, which was originally proposed in computer vision and performed linear interpolations between randomly sampled image pairs to create virtual training data.  \citet{10.1145/3366423.3380144, chen2020mixtext} adapted the idea to textual domains and have applied it to the preliminary task of text classification. However, unlike classifications where each sentence only has one label, sequence labeling tasks such as NER usually involve multiple interrelated labels in a single sentence. As we found in empirical experiments, it is challenging to directly apply such mixup technique to sequence labeling, and improper interpolations may mislead the model. For instance, {\em random sampling} in mixup may inject too much noise by interpolating data points
far away from each other, hence making it fail on sequence labeling. 

To fill this gap, we propose a novel method called \textbf{L}ocal \textbf{A}dditivity based \textbf{D}ata \textbf{A}ugmentation \textbf{(LADA)}, in which we constrain the samples to mixup to be {\em close} to each other. Our
method has two variations: \textbf{Intra-LADA} and \textbf{Inter-LADA}. Intra-LADA interpolates each token's hidden representation with other tokens from the same sentence, which could increase the robustness towards word orderings. Inter-LADA interpolates each token's hidden representation in a sentence with each token from other sentences sampled from a weighted combination of $k$-nearest neighbors sampling and random sampling, the weight of which controls the delicate
trade-off between noise and regularization. 
To further enhance the performance of learning with limited labeled data, we extend LADA to the semi-supervised setting, i.e., \textbf{Semi-LADA}, by designing a novel consistency loss between unlabeled data and its local augmentations.   
We conduct experiments on two NER datasets to demonstrate the effectiveness of our LADA based models over state-of-the-art baselines.

\begin{figure*}[t]
\centering
\includegraphics[width=0.88\textwidth]{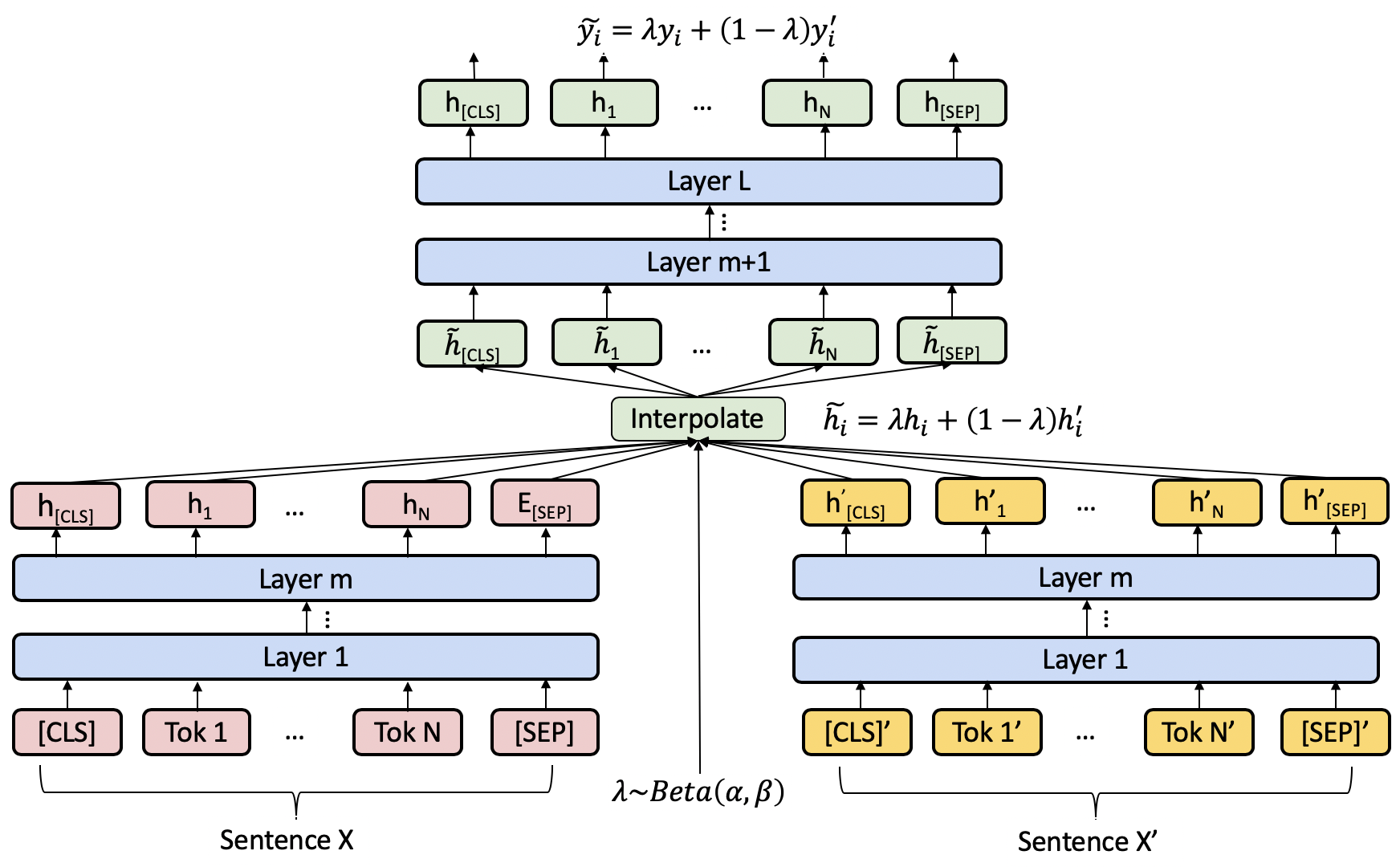}
\caption[]{Overall Architecture of LADA. LADA takes in two sentences, linearly interpolates their hidden states $h_i$ and $h'_i$ at layer $m$ with weight $\lambda$ into $\tilde{h_i}$, and then continues forward passing to get encoded representations $\tilde{h}_i$, which are utilized in downstream tasks where the labels in each task are also mixed with weight $\lambda$.}%
\label{fig:LADA}
\end{figure*}

\section{Background}\label{Sec:background}
\citet{DBLP:journals/corr/abs-1710-09412} proposed a data
augmentation technique called \textsl{mixup}, which trained an image classifier
on linear interpolations of randomly sampled image data. Given a pair
of data points $(\mathbf{x}, \mathbf{y})$ and $(\mathbf{x}^\prime, \mathbf{y}^\prime)$,
where $\mathbf{x}$ denotes an image in raw pixel space, and $\mathbf{y}$ is the
label in a one-hot representation, \textsl{mixup} creates a new sample by interpolating
images and their corresponding labels:
\begin{align*}
    \tilde{\mathbf{x}} &=  \lambda \mathbf{x} + (1-\lambda) \mathbf{x}^\prime, \\
    \tilde{\mathbf{y}} &= \lambda \mathbf{y} + (1-\lambda) \mathbf{y}^\prime,
\end{align*}
where $\lambda$ is drawn from a Beta distribution. \textsl{mixup} trains the neural network for image classification by minimizing the loss on the virtual examples.
In experiments, the pairs of images data points $(\mathbf{x}, \mathbf{y})$ and
$(\tilde{\mathbf{x}}, \tilde{\mathbf{y}})$ are {\em randomly} sampled.
By assuming all the images are mapped to a low dimension manifold through a neural network,
linearly interpolating them creates a virtual vicinity distribution around 
the original data space, thus improving the generalization performance of the
classifier trained on the interpolated samples. 

Prior work like Snippext~\cite{10.1145/3366423.3380144}, MixText~\cite{ chen2020mixtext} and AdvAug \cite{49123} generalized the idea to the textual domain by proposing to interpolate in output space \cite{10.1145/3366423.3380144}, embedding space \cite{49123}, or general hidden space \cite{ chen2020mixtext} of textual data and applied the technique to NLP tasks such as text classifications and machine translations and achieved significant improvements.

\section{Method}
Based on the above interpolation based data augmentation techniques, in Section~\ref{sec:method},
we introduced a \textbf{L}ocal \textbf{A}dditivity based \textbf{D}ata \textbf{A}ugmentation \textbf{(LADA)}
for sequence labeling, where creating augmented samples is much more challenging. 
We continue to describe how to utilize unlabeled data with LADA for semi-supervised NER
in Section~\ref{Sec:semi_lada}.

\subsection{LADA} \label{sec:method}
For a given sentence with $n$ tokens $\mathbf{x} = \{ x_1, ..., x_n\}$, 
denote the corresponding sequence label as $\mathbf{y} = \{y_1, ..., y_n\}$. 
In this paper, we use NER as the working example to introduce our model, in which the labels are the entities types.
We randomly sample a pair of sentences from the corpus, 
$(\mathbf{x}, \mathbf{y})$ and $(\mathbf{x}^\prime, \mathbf{y}^\prime)$, and then
compute the interpolations in the hidden space using a $L$-layer encoder $\mathbf{F}(.; \mathbf{\theta})$.
The hidden representations of $\mathbf{x}$ and $\mathbf{x}^\prime$ up to the $m$-th layer are given by:
\begin{align*}
    \mathbf{h}^l &= \mathbf{F}^l(\mathbf{h}^{l-1}; \mathbf{\theta}), l \in [1,m], \\
    \mathbf{h}'^l &= \mathbf{F}^l(\mathbf{h}'^{l-1}; \mathbf{\theta}), l \in [1,m],
\end{align*}
Here $\mathbf{h}^l = \{h_1, ..., h_n\}$ refer to the hidden representations at the $l$-th layer
and is the concatenation of token representations at all positions. We use $\mathbf{h}^0, \mathbf{h}'^0$ to denote the word embedding of $\mathbf{x}$ and $\mathbf{x}'$ respectively.
At the $m$-th layer, the hidden representations for each token in $\mathbf{x}$ are linearly interpolated 
with each token in $\mathbf{x}'$ by a ratio $\lambda$:
\begin{equation*}
    \tilde{\mathbf{h}}^m = \lambda \mathbf{h}^m + (1-\lambda) \mathbf{h}'^m,
\end{equation*}
where the mixing parameter $\lambda$ is sampled from a Beta distribution, i.e., $\lambda \sim \text{Beta}(\alpha , \alpha)$. Then $\tilde{\mathbf{h}}^m$ is  fed to the upper layers:
\begin{align*}
    \tilde{\mathbf{h}}^l &= \mathbf{F}^l(\tilde{\mathbf{h}}^{l-1}; \theta), l \in [m+1,L].
\end{align*}
$\tilde{\mathbf{h}}^L$ can be treated as the hidden representations of a {\em virtual sample}
$\tilde{\mathbf{x}}$, i.e., $\tilde{\mathbf{h}}^L = \mathbf{F}(\tilde{\mathbf{x}}; \theta)$.

In the meanwhile, their corresponding labels are linearly added with the same ratio: 
\begin{align*}
  \tilde{y}_i =& \lambda y_i + (1-\lambda) y_i^\prime \\
  \tilde{\mathbf{y}} = & \{\tilde{y}_1, ..., \tilde{y}_n\}.
\end{align*}
\noindent
The hidden representations $\tilde{\mathbf{h}}^L$ are then fed into a classifier ${p}(:, \phi)$ and the loss over all positions is minimized to train the model:
\begin{equation} \label{Eq:labeled}
    L=\E_{\mathbf{x}'\sim P_{\text{mix}}(\mathbf{x}' | \mathbf{x})} 
    [\sum_{i=1}^{n} \text{KL} (\tilde{y}_i; p(\tilde{h}_i^L; \mathbf{\phi}))].
\end{equation} 
Here  $P_{\text{mix}}(\mathbf{x}' | \mathbf{x})$ defines the probability of sampling $(\mathbf{x}', \mathbf{y}')$ to mix with $(\mathbf{x}, \mathbf{y})$. The overall diagram is shown in Figure~\ref{fig:LADA}.

Let $\mathbf{S} = \{(\mathbf{x}, \mathbf{y})\}$ be the corpus of data samples, then according to \citet{chen2020mixtext},
\begin{align}
    P_\text{mix}(\mathbf{x}'| \mathbf{x}) = \frac{1}{|\mathbf{S}|}, \quad\quad  (\mathbf{x}', \mathbf{y}') \in \mathbf{S}.
\end{align}
Note that $P_\text{mix}(\mathbf{x}'| \mathbf{x})$ is a uniform distribution that is independent of $\mathbf{x}$.
Even though $\mathbf{x}'$ can be far away from $\mathbf{x}$ in the Euclidean space, they are mapped 
into a low-dimension manifold through a neural network. Interpolating them in the hidden space regularizes
the model to perform linearly in the low-dimensional manifold, hence greatly improves tasks such as 
classification.

\begin{figure}[!t]
\centering
\includegraphics[width=0.98\columnwidth]{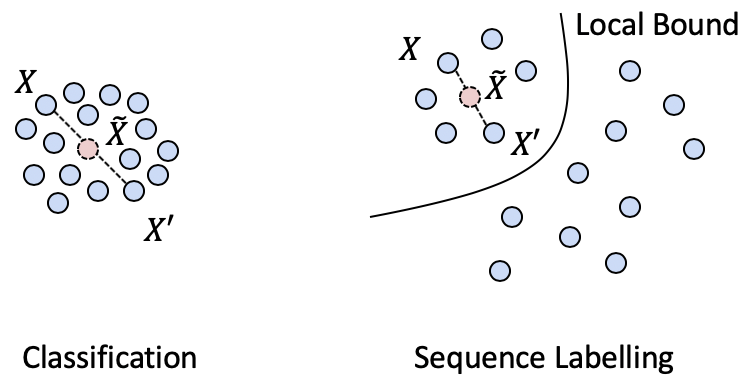}
\caption{ \small Data manifold for sentence classification and sequence labeling.
The dimension of data manifold for sequence labeling is higher
than sentence classification, hence the distance between data samples is larger.
We constraint $\mathbf{x}'$ to be close to $\mathbf{x}$ in
creating interpolated data in LADA.}\label{Fig:data_distance}
\end{figure}

However, we found empirically in experiments that the above {\em random sampling} strategy failed on sequence labeling like NER, leading to worse modeling results than purely supervised learning. 
Intuitively, sequence labeling is more complicated
than sentence classification as it requires learning much more fine-grained information. 
Labeling a token depends on not only the token itself but also the \textsl{context}. We hypothesize
that mixing the sequence $\mathbf{x}$ with $\mathbf{x}^\prime$ changes the context for all tokens and injects too much noise, hence making learning the labels for the tokens challenging. 
In other words, the relative distance between $\mathbf{x}$ and $\mathbf{x}^\prime$ in the manifold mapped by neural networks is further in sequence labeling than sentence classification (demonstrated in Figure~\ref{Fig:data_distance}), which is intuitively understandable as every data point in sentence classification is the pooling over all the tokens in one sentence while every token is a single data point in sequence labeling. Randomly mixing data points far away from each other introduces more noise for sequence labeling. To overcome this problem, we introduce a
\textsl{local} additivity based data augmentation approach with two variations, in which we constrain $\mathbf{x}^\prime$
to be close to $\mathbf{x}$:

\subsection{Intra-LADA}
As stated above, mixing two sequences not only changes the local token representations but also affects the context required to label tokens. To reduce the noises from unrelated sentences, the most direct way is to construct $\mathbf{x}'$ using the same tokens from $\mathbf{x}$ but changing the orders and perform interpolations between them. 

Let $\mathbf{Q} = \text{Permutations}((\mathbf{x}, \mathbf{y}))$ be
the set including all possible permutations of $\mathbf{x}$, then
\begin{align}
    P_\text{Intra}(\mathbf{x}'| \mathbf{x}) = \frac{1}{n!}, \quad\quad (\mathbf{x}',\mathbf{y}') \in \mathbf{Q}.
\end{align}
In this case, each token $x_i$ in $\mathbf{x}$ is actually interpolated with another token $x_j$ in $\mathbf{x}$,
while the context is unaltered. By sampling from $P_\text{Intra}$, we are essentially 
turning sequence level interpolation to token level interpolation, thus greatly reducing
the complexity of the problem.
From another perspective, Intra-LADA generates augmentations with 
different sentence structures using the same word set, 
which could potentially increase the model's robustness towards word orderings.

Intra-LADA restraints the context from changing, which
could be limited in generating diverse augmented data.
To overcome that, we propose Inter-LADA, where we sample a different sentence
from the training set to perform interpolations.

\subsection{Inter-LADA}
Instead of interpolating within one sentence, Intra-LADA samples a different sentence $\mathbf{x}'$ from the training set to interpolate with $\mathbf{x}$. To achieve a trade-off between noise and regularization, we sample $\mathbf{x}'$  through a weighted combination of two strategies: $k$-nearest neighbors ($k$NNs) sampling and random sampling:
\begin{equation}
    P_\text{Inter}(\mathbf{x}'| \mathbf{x})
    = \begin{cases}
    \frac{\mu}{k}, \quad\quad \mathbf{x}' \in \text{Neighbor}_k(\mathbf{x}),\\
    \frac{1-\mu}{|S|}, \quad (\mathbf{x}', \mathbf{y}') \in \mathbf{S},
    \end{cases}
\end{equation}
where $\mu$ is the weight of combining two distributions. 
To get the $k$NNs, we use sentence-BERT \cite{reimers2019sentencebert} 
to map each sentence $\mathbf{x}$ into a hidden space, then collect each sentence's $k$NNs using $l^2$ distance.
For each sentence $\mathbf{x}$, we sample $\mathbf{x}'$ to mix up from the $k$NNs with probability $\mu$ 
and the whole training corpus with a probability $1-\mu$. When $\mathbf{x}'$ is sampled from the whole training corpus, it may be unrelated to $\mathbf{x}$, introducing large noise but also strong regularization on the model.
When $\mathbf{x}'$ is sampled from the $k$NNs, $\mathbf{x}'$ shares similar, albeit different, context with $\mathbf{x}$,
thus achieving good signal to noise ratio. By treating $\mu$ as a hyper-parameter, we can control the delicate trade-off between noise and diversity in regularizing the model.

To examine why sampling sentences from $k$NNs decreases the noise and provides meaningful signals to training, we analyze an example with its $k$NNs in Table~\ref{Tab:knn_examples}:
(1) As it shows, $k$NNs may contain the same entity words as the original sentence, but in different contexts. The entity types in the neighbor sentences are also changed corresponding to contexts. For example, entity \textit{Israel} in the third neighbor becomes an organization when surrounded by \textit{Radio} while it is a location in the original sentence.
(2) Contexts from neighbor sentences can help detect the entities of the same type in a given sentence. For example, \textit{Lebanon} in the second neighbor shares the same type as \textit{Israel} in the original sentence. \textit{Lebanon} can resort to the context of the original sentence to detect its entity type. 
(3) Neighbor sentences may contain the same words but in different forms. For example, the \textit{Israeli} in the first neighbor sentence is a different form of \textit{Israel}, which is miscellaneous while \textit{Israel} is a location in the example sentence. 
Interpolation with such an example can improve models' ability to recognize words of different forms and their corresponding types.

In summary, Inter-LADA can improve both entity learning and context learning by interpolating more diverse data. Note that although we use NER as a working example , LADA can be applied to any sequence labeling models. 
\begin{table*}[!t]
\centering
\begin{tabular}{|c|l|}
\hline
\textbf{Sentence}                            & {\textcolor[HTML]{009901}{Israel}  plays down fears  of war with \textcolor[HTML]{009901}{Syria}.}                                                                                                                                                                                                                                                                                                                                                                       \\ \hline
                             & \begin{tabular}[c]{@{}l@{}}Fears of an \textcolor[HTML]{986536}{Israeli}  operation causes the redistribution of \textcolor[HTML]{986536}{Syrian}   troops locations\\ in \textcolor[HTML]{009901}{Lebanon}  .\end{tabular}                                                                                                                                                                                                                                                                  \\ \cline{2-2} 
                             & \begin{tabular}[c]{@{}l@{}}Parliament Speaker \textcolor[HTML]{9A0000}{Berri}:  \textcolor[HTML]{009901}{Israel}  is preparing for war  against  \textcolor[HTML]{009901}{Syria}  and \textcolor[HTML]{009901}{Lebanon}  .\end{tabular}                                                                                                                                                                                                                                                   \\ \cline{2-2} 
\multirow{-5}{*}{\textbf{Neighbours}} & \begin{tabular}[c]{@{}l@{}} \textcolor[HTML]{9A0000}{Itamar}  \textcolor[HTML]{9A0000}{Rabinovich} , who as \textcolor[HTML]{009901}{Israel}'s ambassador to   \textcolor[HTML]{009901}{Washington}   conducted unfruitful \\negotiations with \textcolor[HTML]{009901}{Syria} , told  \textcolor[HTML]{00009B}{Israel}   \textcolor[HTML]{00009B}{ Radio}  looked like \textcolor[HTML]{009901}{Damascus}   wanted  to  talk \\ rather than fight .\end{tabular} \\ \hline
\end{tabular}
\caption{$k$NNs of an example sentence. Entities in sentences are colored.  \textcolor[HTML]{009901}{Green} means \textcolor[HTML]{009901}{locations} , \textcolor[HTML]{9A0000}{red} means \textcolor[HTML]{9A0000}{persons} , \textcolor[HTML]{00009B}{blue} means \textcolor[HTML]{00009B}{organizations} and \textcolor[HTML]{986536}{yellow} means \textcolor[HTML]{986536}{miscellaneous}.} \label{Tab:knn_examples}
\end{table*}
\subsection{Semi-supervised LADA} \label{Sec:semi_lada}\vspace{-0.05in}
To further improve the performance of learning with less labeled data, we propose a novel LADA-based approach specifically
for unlabeled data. Instead of looking for nearest neighbors, we use back-translation techniques to generate
paraphrases of an unlabeled sentence $\mathbf{x}_u$ in constructing $\mathbf{x'}_u$. 
The paraphrase $\mathbf{x}'_u$, generated via translating $\mathbf{x}_u$ to
an intermediate language and then translating it back, describes the same content as $\mathbf{x}_u$ and should be close to $\mathbf{x}_u$ semantically. However, there is no guarantee that the same entity would appear
in the same position in $\mathbf{x}_u$ and $\mathbf{x}'_u$. In fact, 
the number of tokens in $\mathbf{x}_u$ and $\mathbf{x}'_u$ may not even be the same. 
For instance, for the sentence ``\emph{Rare \textbf{Hendrix} song draft sells for almost \$17,000}'' and its paraphrased sentence ``\emph{A rare \textbf{Hendrix} song design is selling for just under \$17,000}'', although some words are different, the entity \textsl{Hendrix} keeps unchanged, and there are no extra entities added.
That is, both contain one and only one entity 
({\em Hendrix}) of the same type ({\em Person}).
Nevertheless, we empirically found that most paraphrases contain the same number of entities (for any specific type) as
the original sentence.  Inspired by the observation, we propose a new consistency loss to leverage unlabeled data: $\mathbf{x}_u$ and $\mathbf{x}'_u$ should have the same number of entities for any given entity type.

Specifically, for an unlabeled sentence $\mathbf{x}_u$ and its paraphrase $\mathbf{x}'_u$,  
we first guess their token labels with the current model: 
\begin{align*}
    \mathbf{y}_u = p(\mathbf{F}(\mathbf{x}_u; \theta); \phi).
\end{align*}
To avoid predictions being too uniform at the early stage, we sharpen every token prediction ${y}_{u,i} \in \mathbf{y}_u$ with a temperature $T$: 
\begin{align*}
    \hat{y}_{u,i} = \frac{\left({y}_{u,i}\right)^{\frac{1}{T}}}{\left\|\left({y}_{u,i}\right)^{\frac{1}{T}}\right\|_{1}}, 
\end{align*}
where $||.||_1$ denotes the $l1$-norm. We then add the prediction $\hat{y}_{u,i}$ over all tokens in the sentence
to denote its total number of entities for each type:
\begin{align*}
    \hat{y}_{u, \text{ num}} = \sum_{i=1}^n \hat{y}_{u, i}.
\end{align*}
Note that $\hat{y}_{u, \text{num}}$ is the guessed label vector with C-dimensions, where $C$ is the total number of entity types.
The $i$-th element in the $\hat{y}_{u, \text{num}}$ denotes the total number $i$-type entity in the sentence.

During training, we use the same procedure to get the number of entities for original and  each paraphrase sentence (without sharpening). Assume there are $K$ paraphrases, denote the entity number vector for the $k$-the paraphrase as $\hat{y}_{u, \text{num}}^{'k}$.
The consistency objective for unlabeled sentence $\mathbf{x}$ and its paraphrases is:
\begin{equation}
    L_u =  || \hat{y}_{u, \text{num}} - \hat{y}_{u, \text{num}}^{'k} ||^2.
    \label{Eq:Unlabeled}
\end{equation} 
Here we treat $\hat{y}_{u, \text{num}} $ as fixed and back-propagate only through $\hat{y}_{u, \text{num}}'$ to 
train the model.

Taking into account the loss objectives for both labeled and unlabeled data (Equation \ref{Eq:labeled} and Equation \ref{Eq:Unlabeled}), our \textbf{Semi-LADA} training objective is:\vspace{-0.03in} 
\begin{equation*}
    L_{semi} = L + \gamma L_u
\end{equation*}
where $\gamma$ controls the trade-off between the supervised loss term and the unsupervised loss term.

\begin{table}[t!]
\centering
\begin{tabular}{|c|c|c|}
\hline
\textbf{Dataset}        & \textbf{CoNLL} & \textbf{GermEval} \\ \hline
 Train        & 14,987     & 24,000                                                      \\
 Dev          & 3,466      & 2,200                                                       \\
 Test         & 3,684      & 5,100                                                       \\ \hline
 Entity Types & 4          & 12     \\ \hline
Max Sent Length & 142          & 84  \\ \hline
\end{tabular}
\caption{Data statistics and our data split following 
\citet{benikova14:_germev_named_entit_recog_shared_task}.}\label{Tab:Dataset}
\end{table}

\begin{table*}[t]
\centering
\small
\begin{tabular}{|c|c|ccc|ccc|}
\hline
                     &                & \multicolumn{3}{c}{\textbf{CoNLL}}  \vline \vline & \multicolumn{3}{c}{\textbf{GermEval}}   \vline                                                                                                                                     \\  \cline{3-8}
\multirow{-2}{*}{\textbf{Model} }              &  \multirow{-2}{*}{\textbf{Unlabeled data}} & \textbf{5\%}                                                  & \textbf{10\%}                                                 & \textbf{30\%}                                                 & \textbf{5\%}                                                 & \textbf{10\% }                                                & \textbf{30\% }                                               \\ \hline
Flair  \cite{akbik-etal-2019-flair}            & no             & { 79.32}                         & { 86.31}                         & { 89.96}                         & { 66.54}                         & { 67.92}                         & { 74.11}                         \\
Flair + Intra-LADA$\dag$   & no             & -                                                    & -                                                    & -                                                    & -                                                    & -                                                    & -                                                    \\
Flair + Inter-LADA$\dag$    & no             & { 80.84}                         & { 86.33}                         & \textbf{ 90.61}                         & { 67.40}                         & { 70.02}                         & { 74.63}                                                \\ \hline 
BERT     \cite{devlin-etal-2019-bert}          & no             & { 83.28}                         & { 86.85}                         & { 89.28}                         & { 79.64} & { 80.92} & { 82.87} \\
BERT + Intra-LADA$\dag$     & no             & { 83.52} & { 87.54} & { 89.31} & { 79.93}                         & { 81.10}                         & { 82.92}                         \\
BERT + Inter-LADA$\dag$     & no             & { 84.60}                                                & { 87.81}                                                & { 89.68}                                                & { 80.13}                                             & \textbf{ 81.28}                                                & { 83.63}                                        \\ 
BERT + Intra\&Inter-LADA$\dag$     & no             & \textbf{ 84.85}                                                & \textbf{ 87.85}                                                & { 89.87}                                                & {\textbf{  80.17 }}                                                & { 81.23}                                               & {\textbf{ 83.65}}                                                \\\hline \hline
VSL-GG-Hier   \cite{chen-etal-2018-variational}      & yes            & { 83.38}                         & {84.71}                         & { 85.52}                         & -                                                    & -                                                    & -                                                    \\
MT + Noise  \cite{lakshmi-narayan-etal-2019-exploration}    & yes            & { 82.60}                         & { 83.47}                         & { 84.88}                         & -                                                    & -                                                    & -                                                    \\
BERT + Semi-Intra-LADA$\dag$  & yes            & \textbf{ 87.15} &{ 88.70}                                                & { 89.69}                                                & { 80.95} & { 81.52}                         &  { 83.46}                         \\
BERT + Semi-Inter-LADA$\dag$ & yes            & { 86.51}                                                & { 88.53}                                                &{ 90.00}                         & \textbf{ 81.20}                                                & { 81.70}                                                & { 83.53}   \\
BERT + Semi-Intra\&Inter-LADA$\dag$ & yes            & { 86.33}                                                & \textbf{ 88.78}                                                & \textbf{ 90.25}                         & { 81.07}                                                & {\textbf{ 81.77}}                                                & {\textbf { 83.63}}   \\ \hline                      
\end{tabular}

\caption{The F1 scores on CoNLL 2003 and GermEval 2014 training with varying amounts of the labeled training data (5\%, 10\%, and 30\% of the original training set). There were 10,000 unlabeled data for each dataset which was randomly sampled from the original training set. All the results were averaged over 5 runs. $\dag$ denotes our methods.}\label{Tab:Limited_data}
\end{table*}
\section{Experiments}

\subsection{Datasets and Pre-processing}
We performed experiments on two datasets in different languages: \textbf{CoNLL} 2003 \cite{10.3115/1119176.1119195} in English and  \textbf{GermEval} 2014 \cite{benikova14:_germev_named_entit_recog_shared_task} in German. The data statistics are shown in Table~\ref{Tab:Dataset}.  We used the BIO labeling scheme and reported the F1 score. 
In order to make LADA possible in recent transformer-based models like BERT, we assigned labels to special tokens \textrm{[SEP]}, \textrm{[CLS]}, and \textrm{[PAD]}. Since BERT tokenized a token into one or multiple sub-tokens, we not only assigned labels to the first sub-token but also to the remaining sub-tokens following the rules: (1) O word: Oxx$\rightarrow$OOO, (2) I word: Ixx$\rightarrow$III,(3) B word: Bxx$\rightarrow$BII, as such kind of assignment will not harm the performance (ablation study was conducted in Section~\ref{Sec:Ablation}). During the evaluation, we ignored special tokens and non-first sub-tokens for fair comparisons. 

In the fully supervised setting, we followed the standard data splits shown in Table~\ref{Tab:Dataset}. In the semi-supervised setting, we sampled 10,000 sentences in the training set as the unlabeled training data. We adopted FairSeq\footnote{\url{https://github.com/pytorch/fairseq}} to implement the back translation. For CoNLL dataset, we utilized German as the intermediate language and English as the intermediate language for GermEval.

\begin{table*}[t]
\centering
\begin{tabular}{|c|c|c|c|c|}
\hline
                                 \textbf{Model} &\textbf{Setting}                   & \textbf{CoNLL } & \textbf{GermEval}   \\ \hline
Flair  \cite{akbik-etal-2019-flair}    & Token Classification     &92.03      &76.92              \\
Flair + Intra-LADA $\ddag$  & Token Classification  &-          &-                  \\
Flair + Inter-LADA $\ddag$  & Token Classification  &\textbf{92.12}      &\textbf{78.45}              \\ \hline
BERT  \cite{devlin-etal-2019-bert}  & Token Classification            &91.19      &86.12              \\
BERT + Intra-LADA $\ddag$  & Token Classification   &91.22      &86.16              \\
BERT + Inter-LADA $\ddag$  & Token Classification  &\textbf{ 91.83 }     & \textbf{ 86.45 }             \\ \hline 
CVT \cite{Clark_2018}  & Multi-task Learning  &92.60 &-\\
BERT-MRC \cite{li2019unified}  & Reading Comprehension    &93.04      &-                  \\ \hline
\end{tabular}

\caption{The F1 score on CoNLL 2003 and GermEval 2014 training with all the labeled training data. $\ddag$ means incorporating our LADA data augmentation techniques into pre-trained models.}\label{Tab:Full}
\end{table*}

\subsection{Baselines \& Model Settings}
Our LADA can be applied to any models in standard sequence labeling frameworks. In this work, we applied LADA to two state-of-the-art pre-trained models to show the effectiveness:

\begin{itemize}
    \item \textbf{Flair} \cite{akbik-etal-2019-flair}: We used the pre-trained Flair embeddings\footnote{\url{https://github.com/flairNLP/flair}},  and a multi-layer BiLSTM-CRF \cite{DBLP:journals/corr/MaH16} as the encoder to detect the entities. 
    \item \textbf{BERT} \cite{devlin-etal-2019-bert}: We loaded the BERT-base-multilingual-cased\footnote{{\url{https://github.com/huggingface/transformers}}} 
as the encoder and a linear layer to predict token labels.
\end{itemize}

To demonstrate whether our Semi-LADA works with unlabeled data, we compared it with two recent state-of-the-art semi-supervised NER models:

\begin{itemize}
    \item \textbf{VSL-GG-Hier} \cite{chen-etal-2018-variational} introduced a hierarchical latent variables models into semi-supervised NER learning.
    \item  \textbf{MT + Noise} \cite{lakshmi-narayan-etal-2019-exploration} explored different noise strategies including word-dropout, synonym-replace, Gaussian noise and network-dropout in a mean-teacher framework.
\end{itemize}

We also compared our models with another two recent state-of-the-art NER models trained on the whole training set: 
\begin{itemize}
    \item \textbf{CVT} \cite{Clark_2018} performed multi-task learning and made use of 1 Billion Word Language Model Benchmark as the source of unlabeled data.
    \item \textbf{BERT-MRC} \cite{li2019unified} formulated the NER as a machine reading comprehension task instead of a sequence labeling problem. 
\end{itemize}

For \textbf{Intra-LADA}, as it broke the sentence structures, it cannot be applied to Flair that was based on LSTM-CRF. Thus we only combined it with BERT
and only used the labeled data. The mix layer set was \{12\}. For \textbf{Inter-LADA}, we applied it to Flair and BERT 
trained with only the labeled data. The mix layer set was \{8,9,10\}, $k$ in $k$NNs was 3, and 0.5 was a good start point for tuning $\mu$.  \textbf{Semi-LADA} utilized unlabeled data as well. The model was built on BERT. 
The weight $\gamma$ to balance the supervised loss and unsupervised loss was 1.  

\subsection{Main Results} 
We evaluated the baselines and our methods using F1-scores on the test set. 

\paragraph{Utilizing Limited Labeled Data}
We varied the number of labeled data (made use of 5\%, 10\%, 30\% of labeled sentences in each dataset, which were 700, 1400, 4200 in CoNLL and 1200, 2400, 7200 in GermEval) and
the results were shown in Table~\ref{Tab:Limited_data}.  Compared to purely Flair and BERT, applying \textit{Intra-LADA} and \textit{Inter-LADA} consistently boosted performances significantly, indicating the effectiveness of creating augmented training data through local linear interpolations. 
When unlabeled data was introduced, VSL-GG-Hier and MT + Noise performed slightly better than Flair and BERT with 5\% labeled data in CoNLL, but pre-trained models (Flair, BERT) still got higher F1 scores when there were more labeled data. Both kinds of \textit{BERT + Semi-LADA} significantly boosted the F1 scores on CoNLL and GermEval compared to baselines, as Semi-LADA not only utilized LADA on labeled data to avoid overfitting but also combined back translation based data augmentations on unlabeled data for consistent training, which made full use of both labeled data and unlabeled data.

\paragraph{Utilizing All the Labeled data}
Table~\ref{Tab:Full} summarized the experimental results on the full training sets (14,987 on CoNLL 2003 and 24,000 on GermEval 2014). Compared to pre-trained Flair and BERT\footnote{
Note that for the discrepancy between our BERT results and results published in the BERT paper, it has been discussed in the official repo \url{https://github.com/google-research/bert/issues/223}, where the best performance one can replicate on CoNLL was around 91.3 based on the given scripts.
For our experiments, we followed the provided scripts, and kept model settings identical as baselines for fair comparison.}, there were still significant performance gains from utilizing our LADA, which indicated that our proposed data augmentation methods work well even with a large amount of labeled training data (full datasets). We also showed two state-of-the-art NER models' results with different settings, they had better performance mainly due to the multi-task learning with more unlabeled data (CVT) or formulating the NER as reading comprehension problems (BERT + MRC). Note that our LADA was orthogonal to these two models.

\begin{figure}[t!]
\centering
\includegraphics[width=0.95\columnwidth]{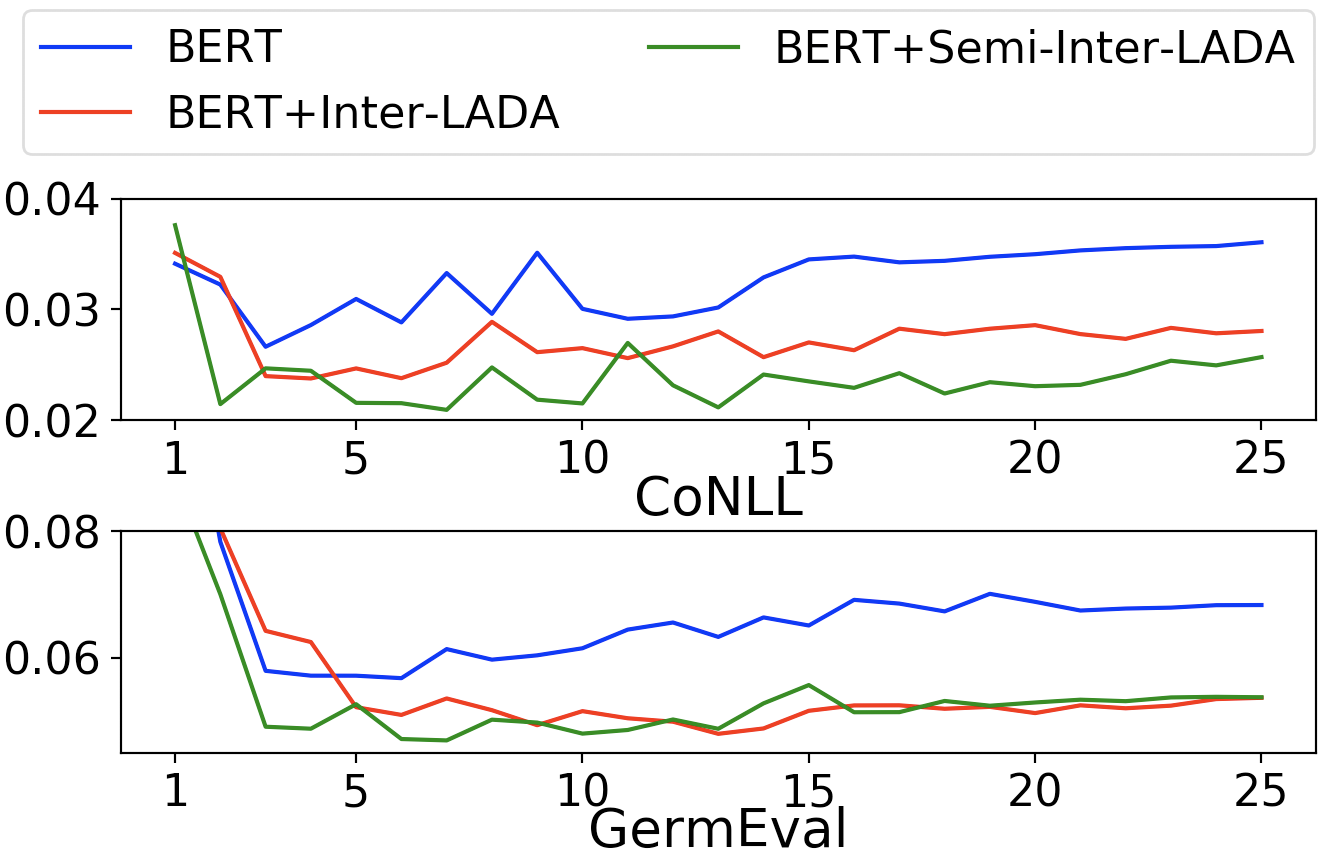}
\caption{Loss (Y axis) on development set, trained with 5\% labeled data, over different epochs (X axis).}\label{Fig:dev_loss}
\end{figure}

\paragraph{Loss on the Development Set}
To illustrate that our LADA could also help the overfitting problem, we plotted the loss on the development set of BERT, \textit{BERT + Inter-LADA} and \textit{BERT + Semi-Inter-LADA} on CoNLL and GermEval training with 5\% labeled data in Figure~\ref{Fig:dev_loss}.  After applying LADA, the loss curve was more stable with training epoch increased, while the loss curve of BERT started increasing after about 10 epochs, indicating that the model might overfit the training data. Such property made LADA a suitable method, especially for semi-supervised learning. 

\paragraph{Combining Intra\&Inter-LADA} We further combined Intra-LADA and Inter-LADA with a ratio $\pi$, i.e. data point would be augmented through Intra-LADA with a probability $\pi$ and Inter-LADA with a probability $1-\pi$. In practice, we set the probability 0.3, and kept the settings for each kind of LADA the same. The results are shown in Table~\ref{Tab:Limited_data}. Through combining two variations, \textit{BERT + Intra\&Inter-LADA} further boosted model performance on both datasets, with an increase of 0.25, 0.04 and 0.19 on CoNLL over \textit{BERT + Inter-LADA} trained with 5\%, 10\% and 30\% labeled data.
We obtained consistent improvement in  semi-supervised settings: \textit{BERT + Semi-Intra\&Inter-LADA} improved over \textit{BERT + Semi-Inter-LADA} trained 
with 5\%, 10\% and 30\% labeled data on GermEval by +0.05, +0.07 and +0.10.
This showed that our Intra-LADA and Inter-LADA can be easily combined by future work to create diverse augmented data to help sequence labeling tasks. 

\subsection{Ablation Study} \label{Sec:Ablation}
\paragraph{Different Sub-token Labeling Strategies} To prove that our pre-processing of labeling sub-tokens for training was reasonable, 
we compared BERT training with different sub-token labeling strategies in Table~\ref{Tab:Tag_subtoken}.``\textbf{None}'' strategy was used in original BERT-Tagger where sub-tokens are ignored during learning. ``\textbf{Real}'' strategy was used in our Inter-LADA where O words' sub-tokens were assigned O (Oxx$\rightarrow$OOO), I and B words' sub-tokens were assigned I (Ixx$\rightarrow$III, Bxx$\rightarrow$BII). ``\textbf{Repeat}'' referred to assigning the original label to each sub-token (Oxx$\rightarrow$OOO, Ixx$\rightarrow$III, Bxx$\rightarrow$BBB). ``O'' means we assigned O to each sub-token (Oxx$\rightarrow$OOO, Ixx$\rightarrow$IOO, Bxx$\rightarrow$BOO). ``Real'' strategy received comparable performances with original BERT models while the other two strategies decreased F1 scores, indicating our strategy mitigated the sub-token labeling issue.

\begin{figure}[t!]
\centering
\includegraphics[width=0.88\columnwidth]{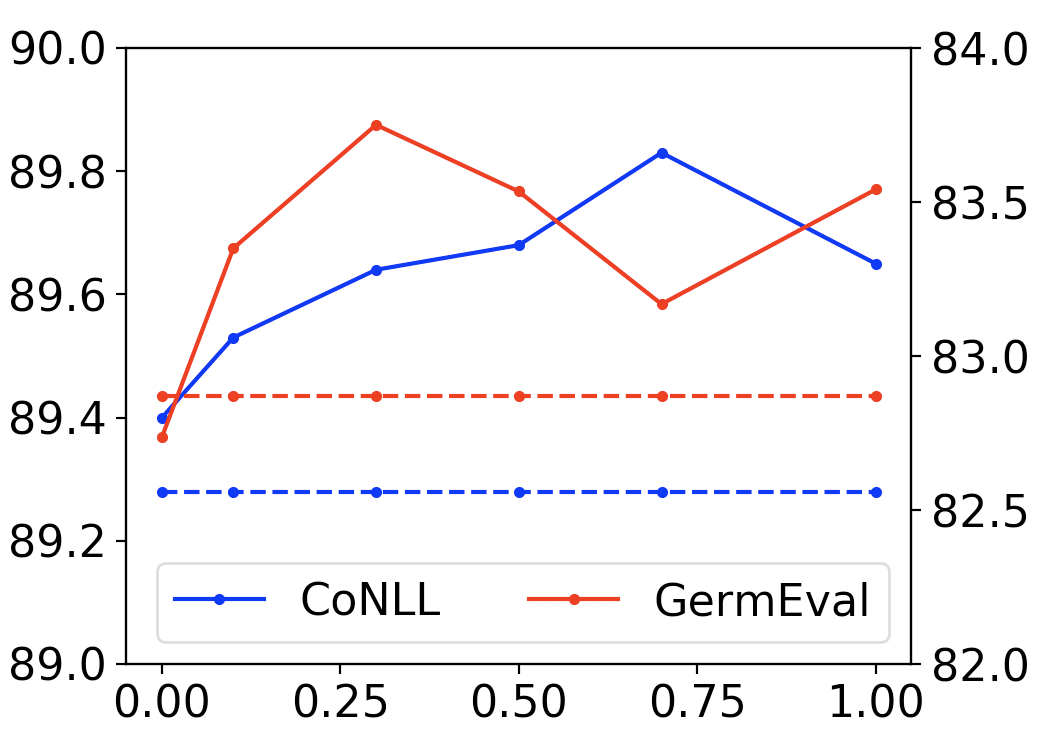}
\caption{F1 score on test set training with 30\%
labeled data with different $\mu$ in BERT + Inter-LADA. The left Y axis is for CoNLL, and the right Y axis is for GermEval. Dashed lines are the F1 scores of BERT model. }\label{Fig:vary_k}
\end{figure}

\paragraph{Influence of $\mu$ in Inter-LADA}  We varied the  $\mu$ in \textit{BERT + Inter-LADA} from 0 to 1 to validate that combining $k$NNs sampling and random sampling in Inter-LADA could achieve the best performance, and the results were plotted in Figure~\ref{Fig:vary_k}. Note that when $\mu = 0$, Inter-LADA only did random sampling and it barely improved over BERT largely due to too much noise from interpolations between unrelated sentences. And when $\mu= 1$, Inter-LADA only did $k$NNs sampling, and it could get a better F1 score over BERT because of providing meaningful signals to training. BERT + Inter-LADA got the best F1 score with $\mu = 0.7$ on CoNLL and $\mu = 0.5$ on GermEval, which indicated the trade-off between noise and diversity ($k$NNs sampling with lower noise and random sampling with higher diversity) was necessary for Inter-LADA.\vspace{-0.05in}

\begin{table}[t]
\centering
\begin{tabular}{|c|c|c|}
\hline
\textbf{Tag Strategy} & \textbf{CoNLL} & \textbf{GermEval}                     \\ \hline
None         & 83.28 & 79.64                        \\ \hline
Real         & 84.15 & 79.59                        \\
Repeat       & 82.67 & 78.27                        \\
O            & 83.13 & 78.48 \\ \hline
\end{tabular}
\caption{F1 scores of BERT on test set with different strategy to tag sub-tokens trained with 5\% labeled data.} \label{Tab:Tag_subtoken}
\end{table}

\section{Related Work} 
\subsection{Named Entity Recognition}
Conditional random fields (CRFs) \cite{LaffertyMP01,SuttonRM04} have been widely used for 
NER, until recently they have been outperformed by neural networks. 
\citet{Hammerton03} and \citet{CollobertWBKKK11} are among the first several studies to model sequence
labeling using neural networks. Specifically \citet{Hammerton03} encoded the input sequence using
a unidirectional LSTM~\cite{hochreiter1997long} while \cite{CollobertWBKKK11} instead used a CNN  with character level embedding to encode sentences. \citet{DBLP:journals/corr/MaH16,LampleBSKD16} proposed LSTM-CRFs to combine neural networks with CRFs that aim to leverage both the representation learning capabilities of neural network and structured loss from CRFs.
Instead of modeling NER as a sequence modeling problem, \citet{li2019unified} converted NER into a reading comprehension task with an input sentence and a query sentence based on the entity types and achieved competitive performance.

\subsection{Semi-supervised Learning for NER}
There has been extensive previous work~\cite{AltunMB05,sogaard-2011-semi,DBLP:journals/jmlr/MannM10} that utilized semi-supervised learning for NER. For instance, \cite{zhang-etal-2017-semi, chen-etal-2018-variational} applied variational autoencoders (VAEs) to semi-supervised sequence labeling; \cite{zhang-etal-2017-semi} proposed to use discrete labeling sequence as latent variables while \cite{chen-etal-2018-variational} used continuous latent variables in their models. 
Recently, contextual representations such as ELMO~\cite{peters-etal-2018-deep} and  BERT~\cite{devlin-etal-2019-bert} trained on a large amount of unlabeled data have been applied to NER and achieved reasonable performances.  Our work is related to research that introduces different data augmentation techniques for NER. For example, \citet{lakshmi-narayan-etal-2019-exploration} applied noise injection and word dropout and obtained a performance boost, \citet{bodapati-etal-2019-robustness} varied the capitalization of words to increase the robustness to capitalization errors, \citet{liu-etal-2019-knowledge}  augmented traditional models with pretraining on external knowledge bases. In contrast, our work can be viewed as data augmentation in the continuous hidden space without external resources.

\subsection{Mixup-based Data Augmentation}
Mixup \cite{DBLP:journals/corr/abs-1710-09412} was originally proposed for image classification \cite{verma2018manifold, Yun_2019} as a data augmentation and regularization method ,
building on which \citet{10.1145/3366423.3380144} proposed to interpolate sentences' encoded representations with augmented sentences by token-substitutions for text classification.
Similarly, \citet{chen2020semisupervised} designed a linguistically informed interpolation of hidden space and demonstrated significant performance increases on several text classification benchmarks.  \citet{49123} performed interpolations at the embedding space in sequence-to-sequence learning for machine translations. Different from these previous studies, we sample sentences based on \textit{local} additivity and utilize mixup for the task of sequence labeling.

\section{Conclusion}
This paper introduced a local additivity based data augmentation (LADA) methods for Named Entity Recognition (NER) with two different interpolation strategies. To utilize unlabeled data, we introduced a novel consistent training objective combined with LADA.  Experiments have been conducted and proved our proposed methods' effectiveness through comparing with several state-of-the-art models on two NER benchmarks. 

\section*{Acknowledgment }
We would like to thank the anonymous reviewers
for their helpful comments, and the members of Georgia Tech SALT group for their feedback. We acknowledge the
support of NVIDIA Corporation with the donation of GPU used for this research. 

\bibliography{emnlp2020}
\bibliographystyle{acl_natbib}

\end{document}